# Sequential Transfer Learning for Plant Disease Recognition Across Perennial Crop Domains Using MobileNetV2-CBAM

**Ropkathaa Choudhury**
Department of Computer Science and Engineering
Adamas University, Kolkata, India
Email: ropkathaachoudhury@gmail.com
ORCID: 0009-0003-2716-4363

**Abstract**

**Purpose**
This study investigates whether a learned disease-recognition representation can be progressively reused across distinct perennial crop domains and how its predictive behaviour changes as target-domain heterogeneity increases.

**Design/methodology/approach**
A lightweight MobileNetV2 framework enhanced with the Convolutional Block Attention Module (CBAM) was evaluated through sequential transfer from Walnut to Cashew, Coffee and Mango, followed by validation on the heterogeneous ABCGMP six-crop benchmark. Each stage used the preceding model representation and was evaluated on an independent test set using complementary classification measures.

**Findings**
Test accuracy increased from 75.76% for the Walnut baseline to 84.85% after CBAM integration, followed by 90.67% on Cashew, 100.00% on Coffee and 99.50% on Mango. The final ABCGMP evaluation achieved 88.44% accuracy, 81.49% Balanced Accuracy and 0.9949 ROC-AUC. The results show that substantial predictive capability was retained through successive crop transfers, while broader multi-crop heterogeneity introduced a measurable performance constraint.

**Originality**
The study provides an empirical characterisation of sequential knowledge reuse across multiple perennial crop domains using a common lightweight attention-enhanced representation, extending evaluation beyond isolated source-to-target transfer.

**Keywords:** transfer learning; perennial crops; plant disease recognition; MobileNetV2; CBAM; sequential transfer; cross-domain adaptation; lightweight deep learning

## 1. Introduction

Plant disease recognition across perennial crops presents a transferability challenge because disease-related visual features learned from one crop may not remain equally effective when applied to another (1). Although crop-specific models can be developed independently, this approach requires separate development and validation for each domain and does not reveal whether learned representations can be reused across crops (1,2). Perennial crops provide a suitable setting for examining this issue because they share a common agricultural context while differing in morphology, disease characteristics and visual appearance. The specific problem addressed in this study is therefore how a learned disease representation behaves when transferred sequentially across distinct perennial crop domains.

Cross-crop transfer is valuable because agricultural disease-recognition systems may need to accommodate additional crop domains without requiring an entirely new learning process for each one (3). At the same time, crops differ in morphology, symptom expression and image characteristics, so knowledge that is useful in one domain may not retain the same value in another (1). Examining these changes across successive transfers therefore provides a way to distinguish reusable disease-related representations from performance that depends more strongly on the target domain.

Deep-learning and transfer-learning approaches have demonstrated effective plant disease classification across a range of datasets and target conditions (2,4–6). However, existing evaluations commonly focus on individual datasets, limited-data transfer or isolated source-to-target adaptation (2,5,7,8). Such evaluations establish whether a model can adapt to a particular target, but provide limited insight into how its learned representation behaves when adaptation is repeated across successive crop domains. This leaves an important empirical question: to what extent can a shared disease-recognition representation be retained and adapted as the target crop and task characteristics change progressively?

This study investigates sequential cross-perennial transfer using a common MobileNetV2 framework enhanced with CBAM. The learned representation is transferred progressively from Walnut to Cashew, Coffee and Mango, followed by evaluation on the heterogeneous ABCGMP six-crop benchmark. Stage-wise evaluation is used to examine how predictive behaviour changes as crop identity, class structure and visual diversity vary across successive target domains. The contribution is therefore an empirical characterisation of sequential knowledge reuse using a lightweight attention-enhanced framework, providing a reproducible basis for assessing transfer behaviour beyond isolated source-to-target experiments.

## 2. Related works

Transfer learning has become an established approach for plant disease recognition because pretrained representations can be adapted to new classification tasks with fewer task-specific resources. Chen et al. demonstrated deep transfer learning for image-based plant disease identification, while Xu et al. investigated transfer learning across multiple plant-disease datasets under limited target data (4,5). Subsequent work has examined the transferability of learned features and the influence of parameter initialisation, highlighting that the usefulness of transferred representations depends on how source and target tasks are related (1,9). These studies establish the basis for knowledge reuse, but leave the cumulative effect of repeated adaptation across successive crop domains insufficiently characterised.

Lightweight architectures have received increasing attention for agricultural disease recognition because practical deployment may impose computational and resource constraints. Khan et al. demonstrated plant disease detection for edge-computing devices, while Attri et al. investigated TinyML-based recognition for apple and mango leaves. Ugwah and Abolghasemi further reported an optimised MobileNet-based approach for lightweight plant leaf disease detection, and Duhan et al. developed a lightweight MobileNetV2-based model for plant disease classification (10–13). Related work has also questioned the need for increasingly deep CNN architectures for plant disease identification (14). Collectively, these studies establish the value of computationally efficient representations, but do not address whether such representations can be retained and adapted through successive perennial crop domains.

Attention mechanisms provide a complementary means of refining visual representations by emphasising informative feature responses. Woo et al. introduced CBAM through sequential channel and spatial attention, while Mahapatra et al. evaluated an attention-based CNN for intra-dataset and cross-dataset plant disease classification (15,16). Multi-crop transfer-learning studies have also combined attention or related architectural enhancements to improve disease recognition across crop domains (17). These studies support attention-based feature refinement as a useful component of plant disease classifiers, but provide limited evidence on how an attention-enhanced representation behaves under sequential transfer across multiple crops.

Cross-domain studies have demonstrated that changes between source and target distributions can substantially affect plant disease recognition. Fuentes et al. investigated open-set and cross-domain adaptation for tomato disease recognition, while Yang et al. examined cross-domain few-shot learning under laboratory-to-field conditions (7,8). Tunio et al. addressed domain discrepancy through Wasserstein-based adaptation with transformer-fused convolutional features, and Xiang et al. reported substantial reliability degradation under controlled-to-field shift (18,19). Broader studies of domain generalisation and domain-adversarial learning further examine strategies for improving robustness when source and target distributions differ (20,21). Lightweight cross-domain recognition has also been explored through target-domain statistical calibration (22). These studies establish the importance of domain shift in plant disease recognition, while the cumulative effect of sequential transfer across multiple perennial crop domains remains insufficiently characterised.

Multi-dataset evaluation has broadened assessment beyond isolated crop benchmarks. Krishna et al. investigated plant leaf disease detection across multiple datasets, while Richter and Kim benchmarked transfer-learning performance across open plant-disease datasets (2,6). Anton et al. developed a transfer-learning framework for multi-crop leaf disease classification, and the ABCGMP dataset provides a heterogeneous benchmark spanning multiple crops and disease categories (17,23). Lightweight and deployment-oriented studies have further examined MobileNet- and TinyML-based disease recognition, AIoT architectures and agricultural deployment (24,25). Lightweight MobileNet- and TinyML-based disease recognition has also been investigated (11–13,26). These studies broaden the evaluation and deployment context of agricultural disease recognition, but do not establish how a common learned representation evolves when transferred sequentially across multiple crop domains.

Taken together, the literature establishes strong foundations for plant disease recognition through transfer learning, lightweight architectures, attention-based refinement, multi-crop learning and cross-domain adaptation (2,4–8,10,16–18**)**. Recent work also demonstrates the application of these approaches to constrained agricultural settings and crop-specific transfer tasks (1,3,9,12–14,19–22,25–29). Studies addressing crop-specific disease recognition and transfer-learning applications further demonstrate the breadth of existing approaches, including limited-data CNN classification and transfer-learning-based Cashew disease analysis (28,29). However, existing evaluations remain largely centred on individual datasets, isolated source-to-target adaptation, or broader domain generalisation settings. Limited empirical evidence is available on how a common lightweight attention-enhanced representation behaves when knowledge is transferred sequentially across multiple crop domains, especially for perennial crops. This study addresses this gap by systematically characterising that behaviour across successive crop transfers and a final heterogeneous multi-crop benchmark.

## 3. Research Methodology

The experimental methodology was designed to examine the sequential transfer of a lightweight disease-classification representation across distinct perennial-crop domains. The same MobileNetV2-CBAM framework was retained throughout the experiments, while the target-specific classifier and selected higher-level parameters were adapted to each dataset. The procedure comprised dataset preparation, model construction, sequential transfer, fine-tuning, training and stage-wise evaluation.

### *3.1 Datasets*

Five publicly available datasets were used in a fixed experimental sequence. Walnut provided the initial 18-class source task, followed by Cashew (2 classes), Coffee (3 classes), Mango (8 classes) and the ABCGMP six-crop benchmark (24 classes). The datasets were assigned distinct experimental roles so that transfer could be examined under progressively different class structures and visual conditions, as detailed in Table I (23,30–33).

Table I. Datasets used in the sequential transfer-learning framework

| Dataset | Crop(s) | Classes | Total Images | Role |
|---|---|---|---|---|
| Walnut Leaf Dataset | Walnut | 18 | 1751 | Source-domain baseline |
| Cashew Leaf Dataset | Cashew | 2 | 500 | First transfer stage |
| Coffee Disease Dataset | Coffee | 3 | 22,588 | Second transfer stage |
| MangoLeafBD Dataset | Mango | 8 | 4,000 | Third transfer stage |
| ABCGMP Fruit and Leaf Disease Six-Crop Dataset | Apple, Banana, Citrus, Guava, Mango, Papaya | 24 | 9,858 | Final cross-domain validation |

The transfer order was defined as ImageNet → Walnut → Cashew → Coffee → Mango → ABCGMP. The Walnut model therefore formed the initial task-specific representation, whereas each subsequent stage used the parameters obtained from the preceding stage.

### *3.2 Image Preparation*

Images were resized to 224 × 224 pixels and normalised by 1/255 rescaling. To increase tolerance to ordinary imaging variation, training data were augmented using rotation, horizontal and vertical flipping, brightness adjustment and zoom. These operations were applied before model optimisation, while the original class structure was retained for evaluation.

### *3.3 Classification Framework*

MobileNetV2 was used as the convolutional backbone and initialised with ImageNet weights for the Walnut stage (34). CBAM was incorporated as the attention module to refine the extracted feature representation (15). The resulting architecture was retained across the transfer stages, with the final classification layer replaced whenever the target dataset required a different number of output classes, as illustrated in Figure 1.

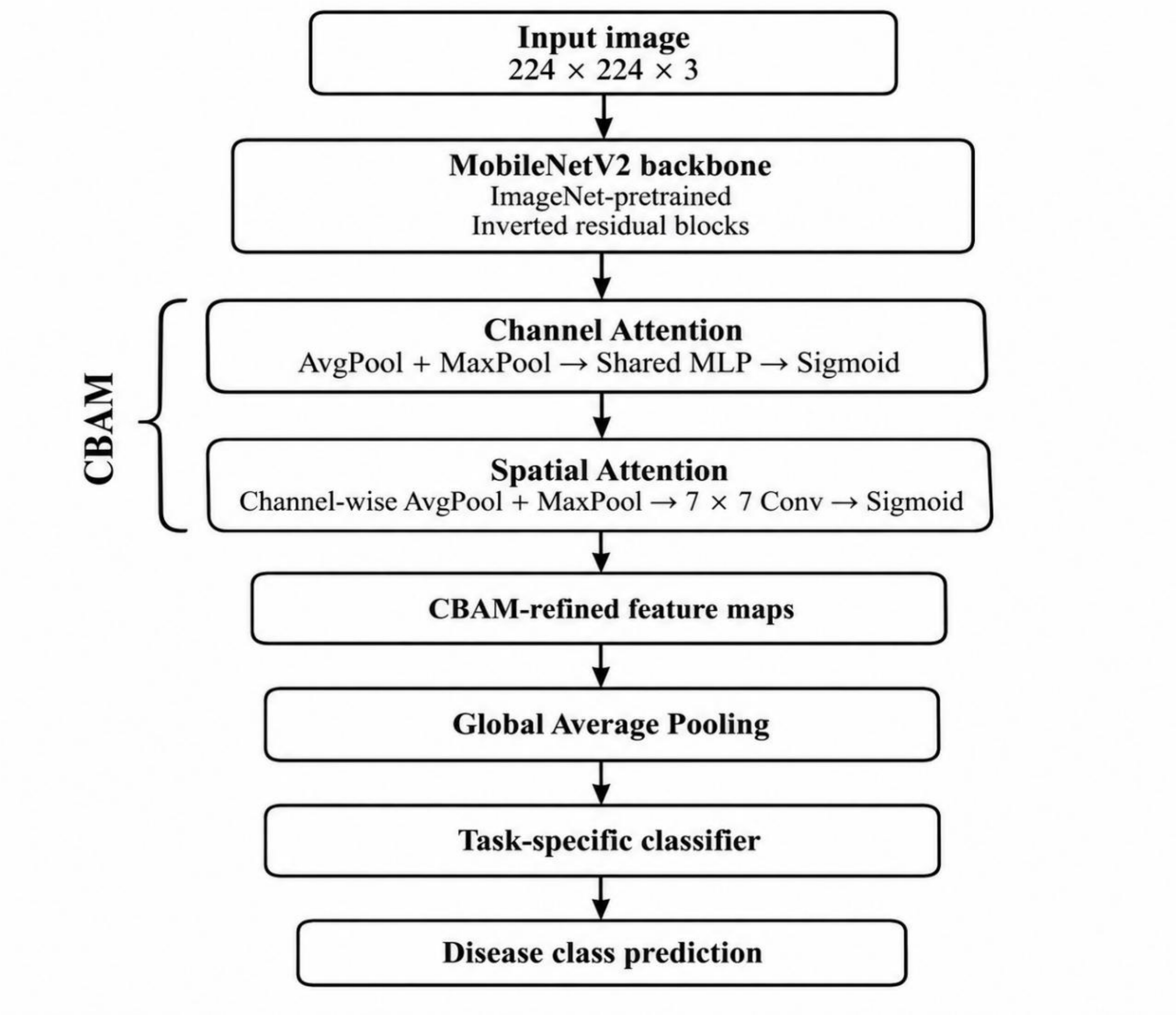

Figure 1. MobileNetV2-CBAM classification architecture.

CBAM refines the extracted feature representation through sequential channel and spatial attention (15).

The channel attention module generates channel-wise attention weights from global average- and maximum-pooled representations, which are then processed through a shared multilayer perceptron (15).

$$M_{c(F)} = \sigma\left(MLP\big(AvgPool(F)\big) + MLP\big(MaxPool(F)\big)\right) \tag{1}$$

Where $F$ is the input feature map, $AvgPool$ and $MaxPool$ denote global average and maximum pooling operations, respectively, $MLP$ represents the shared multilayer perceptron, and $\sigma$ denotes the sigmoid activation function. The resulting $M_{c(F)}$ represents the channel-attention map.

The spatial attention module then identifies informative spatial regions from the channel-refined feature map using channel-wise average and maximum pooling followed by a convolutional operation (15).

$$M_{s(F)} = \sigma\left(f^{7x7}\big(\big[AvgPool_{c(F)}; MaxPool_{c(F)}\big]\big)\right) \tag{2}$$

Where F is the input feature map, $f^{7x7}$ denotes a 7 × 7 convolution, $AvgPool_{c(F)}$ and $MaxPool_{c(F)}$ denote channel-wise average and maximum pooling, respectively; [; ] denotes channel-wise concatenation, and σ denotes the sigmoid activation function.

$$F' = M_c(F) \odot F \tag{3}$$

where F′ denotes the channel-refined feature map and ⨀ denotes element-wise multiplication.

$$F'' = M_s(F') \odot F' \tag{4}$$

where F″ denotes the final CBAM-refined feature map after spatial attention and ⨀ denotes element-wise multiplication.

*3.4 Sequential Transfer and Fine-Tuning*

Following completion of a source-domain experiment, its learned parameters were transferred directly to the next dataset rather than reinitialised from ImageNet. This progressive reuse of learned parameters follows the general principle that the usefulness of transferred features depends on the relationship between the source and target tasks (1,9). During adaptation, MobileNetV2 Blocks 1–15 remained frozen, while Blocks 16–17, CBAM and the target classification head were fine-tuned. The classification head was reinitialised for every target domain to match its class count.

$$\theta_t = FineTune(\theta_{(t-1)}, D_t), \ \ t = 1, \dots, T \tag{5}$$

where $\theta_t$ denotes the model parameters after adaptation to the t-th target domain, $\theta_{(t-1)}$ denotes the parameters inherited from the preceding transfer stage, and $D_t$ represents the corresponding target-domain dataset.

This configuration preserved the lower-level representation while allowing the higher-level feature space to adapt to target-domain characteristics.

$$\theta_{t^*} = arg\ min_{(\theta_{t^{FT}})}\ L(D_t; \theta_{t^{FT}}, \theta^F) \tag{6}$$

where $\theta_{t^*}$ denotes the optimised parameters for the target domain, $\theta_{t^{FT}}$ represents the trainable parameters during fine-tuning, $\theta^F$ denotes the frozen parameters, $D_t$ is the target-domain dataset, and L denotes the training loss.

The resulting model was then used as the source model for the next stage, producing the complete sequential transfer pathway.

$$\theta_{t^{(0)}} = \theta_{(t-1)^*} \tag{7}$$

where $\theta_{t^{(0)}}$ denotes the initial parameters of the current target-domain model and $\theta_{(t-1)^*}$ denotes the optimised parameters obtained from the preceding transfer-learning stage.

The complete sequential cross-perennial transfer-learning pathway is shown in Figure 2. The partial fine-tuning strategy applied across the transfer stages is shown in Figure 3.

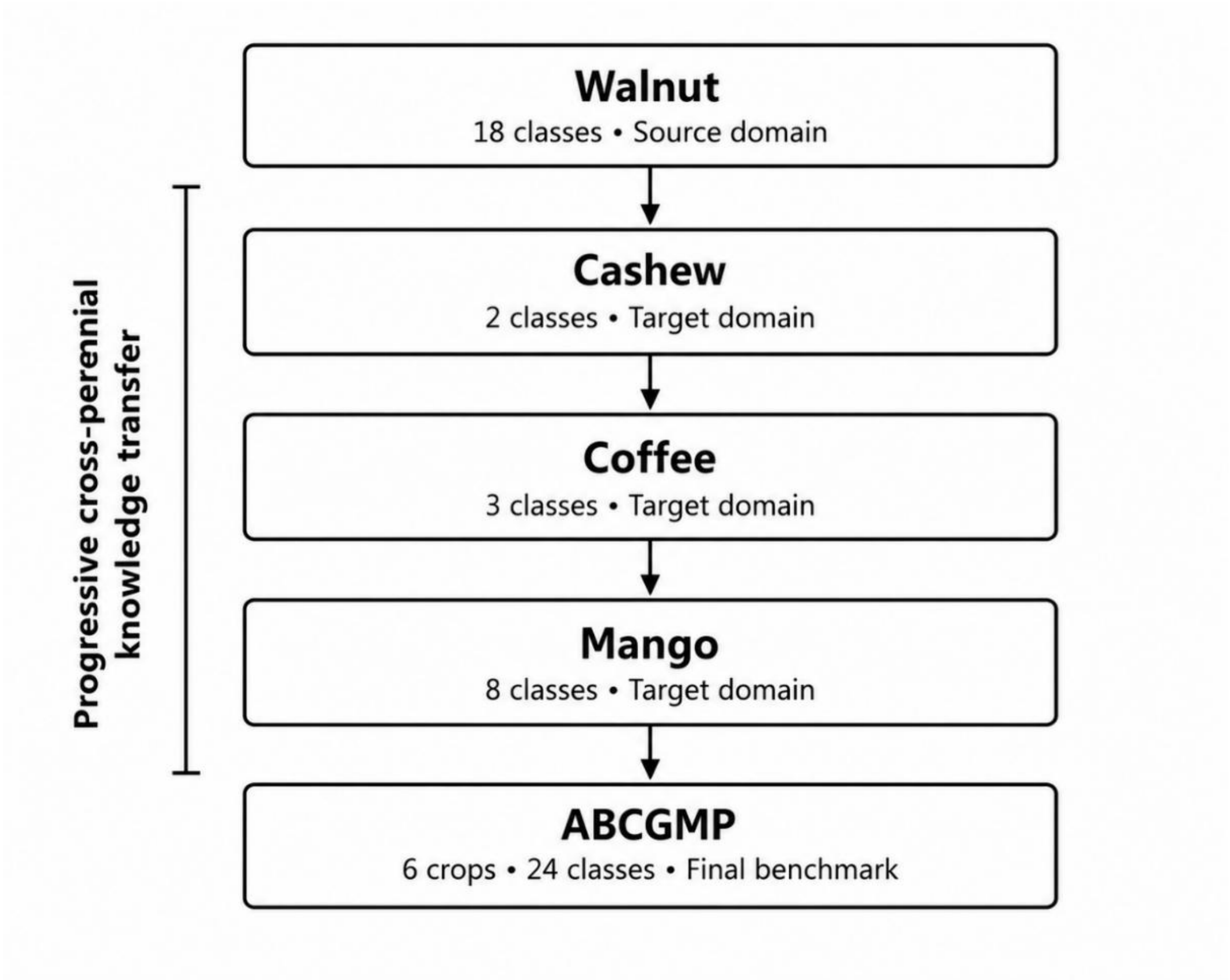

Figure 2. Sequential cross-perennial transfer-learning pathway.

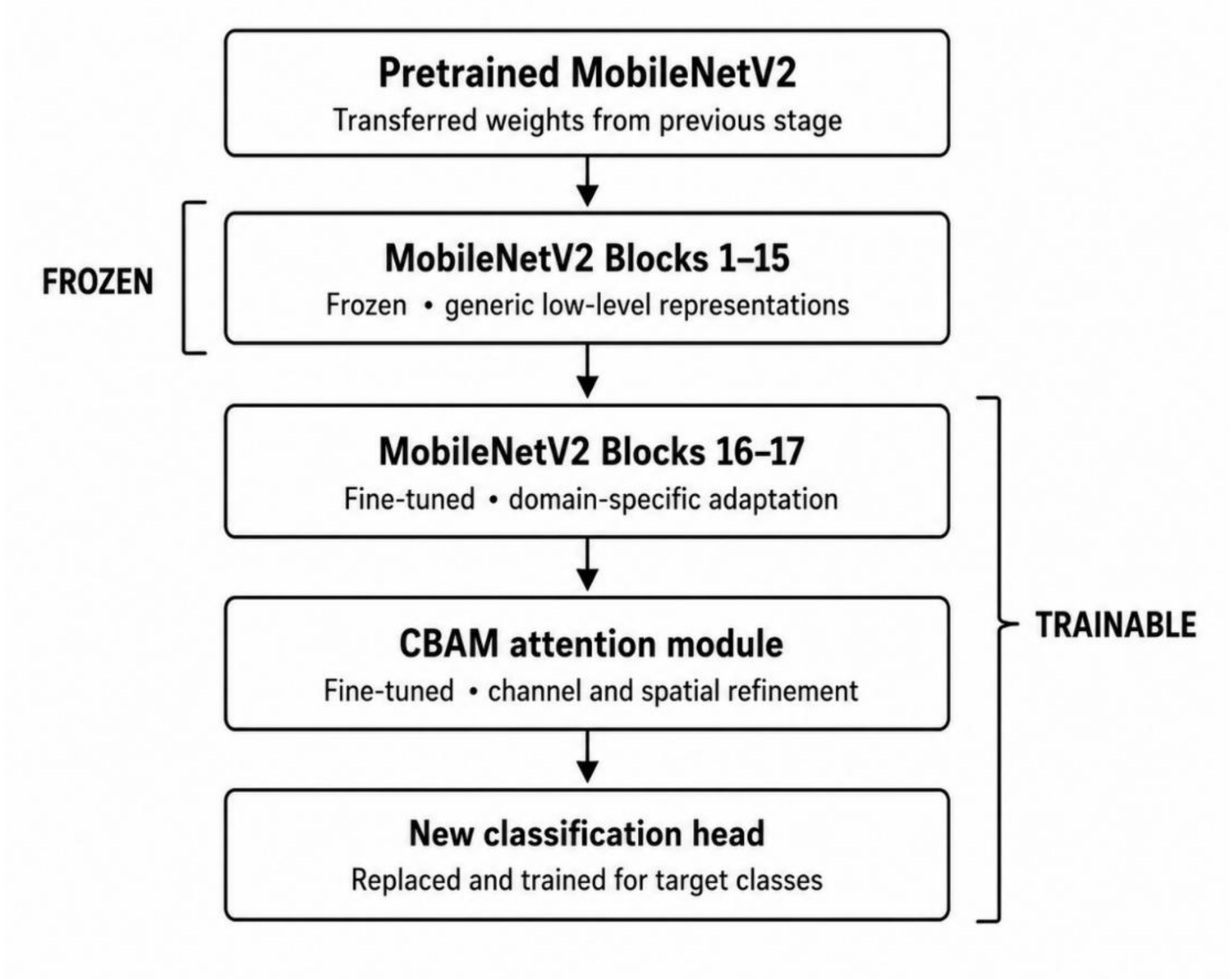

Figure 3. Partial fine-tuning strategy across transfer stages.

*3.5 Training Configuration*

The Walnut baseline was trained with Adam at a learning rate of $5 \times 10^{-4}$, whereas subsequent transfer stages used AdamW at $1 \times 10^{-4}$. ReduceLROnPlateau, early stopping and validation-based checkpointing were applied throughout the experiments. Batch size was 32 for Walnut, Cashew and Coffee and 64 for Mango; maximum training duration ranged from approximately 5 to 20 epochs. Cross-Entropy Loss was used for the standard stages, while Weighted Cross-Entropy was adopted for the class-imbalanced ABCGMP benchmark (35). The stage-specific training configuration is summarised in Table II.

Table II. Sequential transfer-learning configuration

| Stage | Transfer Path | Classes | Trainable Components | Optimiser | Batch Size | Epochs |
|---|---|---|---|---|---|---|
| 1 | ImageNet → Walnut | 18 | Final MobileNetV2 block + classifier | Adam | 32 | 20 |
| 2 | Walnut → Cashew | 2 | MobileNetV2 Blocks 16–17 + CBAM + classifier | AdamW | 32 | 25 |
| 3 | Cashew → Coffee | 3 | MobileNetV2 Blocks 16–17 + CBAM + classifier | AdamW | 32 | 5 |
| 4 | Coffee → Mango | 8 | MobileNetV2 Blocks 16–17 + CBAM + classifier | AdamW | 64 | 10 |
| 5 | Mango → ABCGMP | 24 | Higher MobileNetV2 blocks + CBAM + classifier | AdamW | 32 | 20 |

Cross-Entropy Loss was used as the classification objective for the standard transfer-learning stages.

$$L_{CE} = -\sum_{c=1}^{C} y_c \, log(p_c) \quad (8)$$

where C denotes the number of classes, $y_c$ is the ground-truth indicator for class c, and $p_c$ is the predicted probability for class c.

For the class-imbalanced ABCGMP benchmark, Weighted Cross-Entropy Loss was used to give greater consideration to under-represented classes during optimisation.

$$L_{WCE} = -\sum_{c=1}^{C} w_c \, y_c \, log(p_c) \quad (9)$$

where $w_c$ denotes the weight assigned to class c to compensate for class imbalance, while $y_c$ and $p_c$ denote the ground-truth indicator and predicted probability for class c, respectively.

### *3.6 Evaluation Protocol*

Each transfer stage was evaluated independently on its unseen test set using the best validation checkpoint. Accuracy, Precision, Recall and F1-score were used as the principal classification measures, complemented by Balanced Accuracy, Matthews Correlation Coefficient (MCC), Cohen's Kappa and ROC-AUC to provide a broader assessment of classification behaviour (36–40). The same evaluation protocol was retained across stages so that changes in predictive behaviour could be compared along the transfer sequence.

Accuracy was calculated as the proportion of correctly classified samples in the evaluated test set.

$$Accuracy = \frac{N_{(correct)}}{N_{(total)}} \quad (10)$$

where $N_{(correct)}$ denotes the number of correctly classified samples, and $N_{(total)}$ denotes the total number of evaluated samples.

F1-score was used to combine Precision and Recall into a single harmonic-mean measure.

$$F1 = \frac{2 \times Precision \times Recall}{Precision + Recall} \quad (11)$$

where Precision and Recall denote the precision and recall values, respectively.

The stage-wise evaluation workflow is summarised in Figure 4.

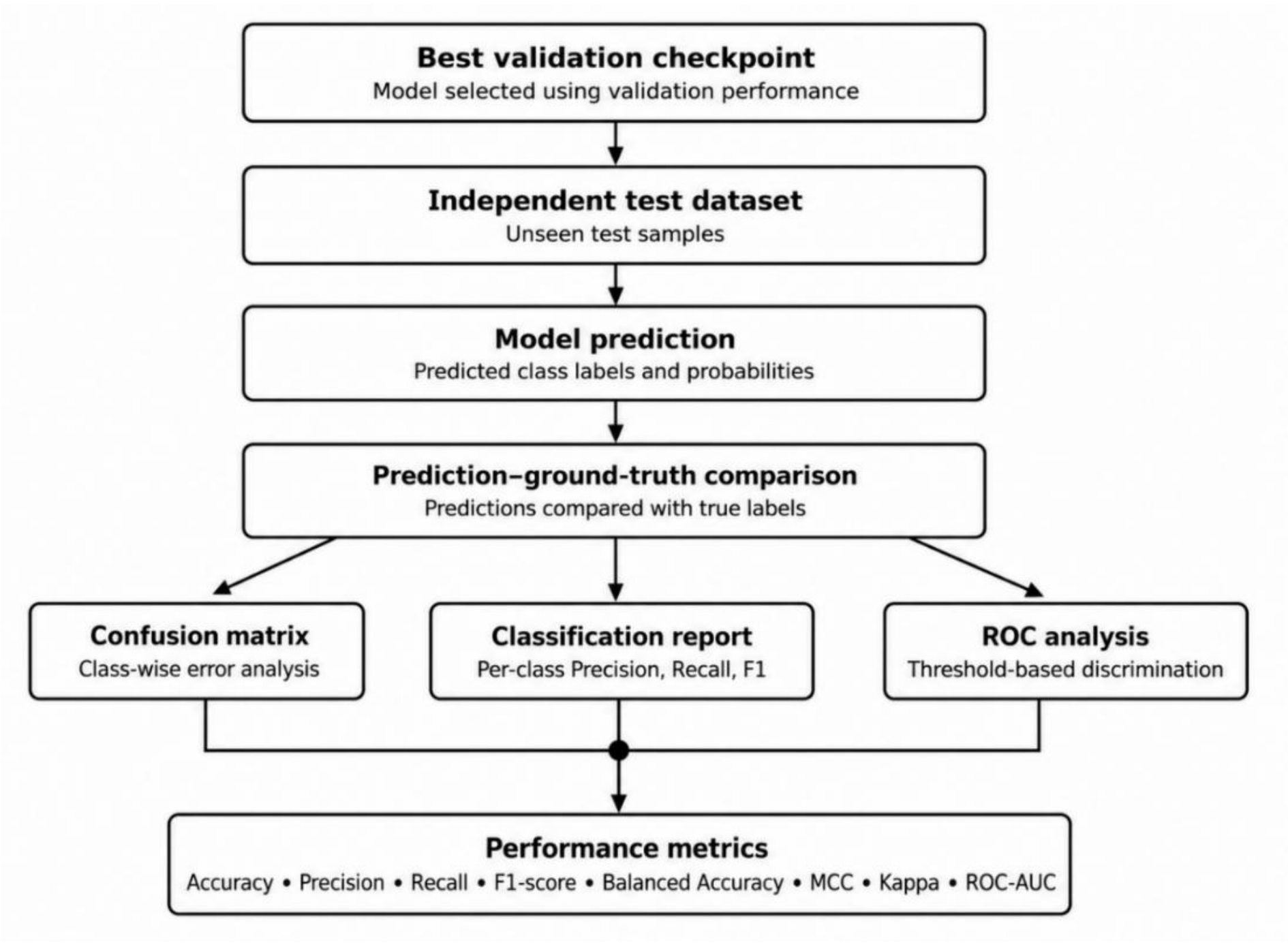

Figure 4. Evaluation workflow for stage-wise performance assessment.

## 4. Results

The experiments progressed from the Walnut source-domain model through successive perennial-crop adaptations, followed by evaluation on a heterogeneous multi-crop benchmark. This sequence enabled the learned representation to be examined at each transfer stage as the target task changed in crop, class structure and visual diversity. The resulting stage-wise classification performance is presented in Table III.

Table III. Core classification performance across the experimental stages

| Stage | Experiment | Accuracy (%) | Precision (%) | Recall (%) | F1-score (%) |
|---|---|---|---|---|---|
| 1 | Walnut baseline | 75.76 | 77.80 | 75.76 | 76.02 |
| 2 | Walnut + CBAM | 84.85 | 85.43 | 84.85 | 84.88 |
| 3 | Walnut → Cashew | 90.67 | 90.98 | 90.67 | 90.68 |
| 4 | Cashew → Coffee | 100.00 | 100.00 | 100.00 | 100.00 |
| 5 | Coffee → Mango | 99.50 | 99.51 | 99.50 | 99.50 |
| 6 | Mango → ABCGMP | 88.44 | 89.95 | 88.44 | 88.78 |

### *4.1 Baseline and Attention Enhancement*

The ImageNet-initialized MobileNetV2 model established the Walnut source representation with a test accuracy of 75.76%. Incorporating CBAM increased the test accuracy to 84.85%, corresponding to a 9.09-percentage-point improvement, with Precision, Recall and F1-score of 85.43%, 84.85% and 84.88%, respectively. The CBAM configuration reached 82.44% validation accuracy and 99.10% training accuracy, while 640,116 of 2,451,828 total parameters were trainable. The resulting attention-enhanced representation provided the starting model for the subsequent cross-crop experiments. The corresponding parameter statistics are provided in Table IV.

Table IV. Model parameter statistics

| Model | Total Parameters | Trainable Parameters | Frozen Parameters |
|---|---|---|---|
| MobileNetV2 baseline | 2,246,930 | 23,058 | 2,223,872 |
| MobileNetV2 + CBAM | 2,451,828 | 640,116 | 1,811,712 |

*4.2 Initial Cross-Crop Transfer*

Transfer of the Walnut representation to Cashew produced 90.67% test accuracy, with Precision of 90.98%, Recall of 90.67% and F1-score of 90.68%. Balanced Accuracy reached 90.89%, while MCC, Cohen's Kappa and ROC-AUC were 0.8161, 0.8135 and 0.9536, respectively. The result indicates that substantial predictive performance was retained after the source representation was adapted to a different crop and target-class structure, providing the next model in the sequential transfer chain.

*4.3 Successive Cross-Crop Transfer*

The Cashew model was subsequently adapted to the three-class Coffee dataset. The resulting model achieved 100.00% Accuracy, Precision, Recall, F1-score, Balanced Accuracy, MCC, Cohen's Kappa and ROC-AUC, with all 3,389 reported test images correctly classified. The result indicates that the adapted representation maintained strong predictive performance after a second cross-crop transition, extending the observed transfer behaviour across successive perennial domains.

*4.4 Higher-Complexity Target Transfer*

The Coffee model was transferred to the eight-class MangoLeafBD dataset (30), where test accuracy reached 99.50%, with Precision of 99.51%, Recall of 99.50% and F1-score of 99.50%. Balanced Accuracy was 99.55%, while MCC and Cohen's Kappa were both 0.9943, and ROC-AUC reached 0.999987. Only three of 600 test images were misclassified, corresponding to an error rate of 0.5%. The sustained performance after successive adaptation indicates that the transferred representation remained effective as the target domain increased in class complexity.

*4.5 Heterogeneous Multi-Crop Validation*

The final transfer was evaluated on the ABCGMP six-crop benchmark (23), comprising 9,858 images across 24 classes and six crops with both fruit and leaf samples. This setting introduced substantially greater variation in morphology, disease appearance, plant organ and class distribution than the preceding single-crop targets. The model achieved 88.44% Accuracy, 89.95% Precision, 88.44% Recall and 88.78% F1-score, with Balanced Accuracy of 81.49%, MCC of 0.8715, Cohen's Kappa of 0.8709 and ROC-AUC of 0.9949. The lower overall accuracy compared with the preceding crop-specific stages indicates the greater difficulty introduced by the heterogeneous final target, while the high ROC-AUC shows that strong class-separation capability was retained across the 24-class setting.

Table V. Advanced evaluation metrics for sequential transfer learning

| Stage | Transfer Experiment | Balanced Accuracy (%) | MCC | Cohen's Kappa | ROC-AUC |
|---|---|---|---|---|---|
| 1 | Walnut → Cashew | 90.89 | 0.8161 | 0.8135 | 0.9536 |
| 2 | Cashew → Coffee | 100.00 | 1.0000 | 1.0000 | 1.0000 |
| 3 | Coffee → Mango | 99.55 | 0.9943 | 0.9943 | 0.999987 |
| 4 | Mango → ABCGMP | 81.49 | 0.8715 | 0.8709 | 0.9949 |

### *4.6 Overall Findings*

The stage-wise results show that the transferred representation maintained strong predictive performance through successive perennial-crop adaptations, while the final heterogeneous benchmark introduced a measurable performance constraint. The progression from crop-specific targets to the six-crop ABCGMP setting therefore provides empirical evidence of how predictive behaviour changes as crop identity, class structure, plant organ and visual diversity become increasingly heterogeneous. The progression of test accuracy across the experimental stages is illustrated in Figure 5.

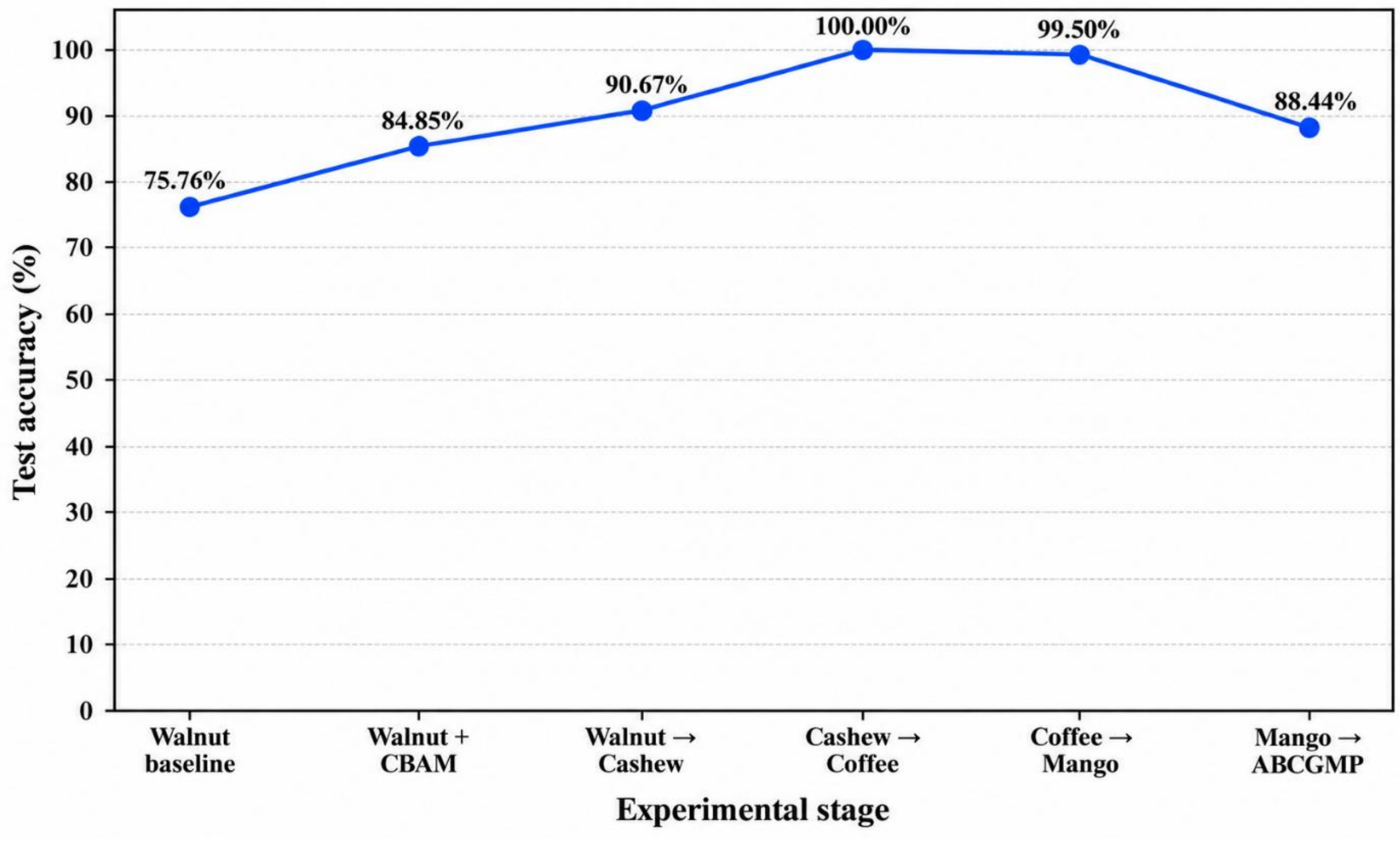

Figure 5. Progressive test accuracy across the experimental stages.

## 5. Conclusion

The study examined whether a learned disease-recognition representation could be progressively reused across distinct perennial crop domains. Starting from the Walnut source model, sequential adaptation to Cashew, Coffee and Mango produced test accuracies of 90.67%, 100.00% and 99.50%, respectively, while the final heterogeneous ABCGMP evaluation achieved 88.44% accuracy. The results show that substantial predictive capability was retained through successive crop transfers, although broader multi-crop heterogeneity introduced a measurable performance constraint.

The principal contribution of this study is the empirical characterisation of sequential knowledge reuse across multiple perennial crop domains using a common lightweight MobileNetV2-CBAM representation. Rather than introducing a new learning architecture, the study provides a reproducible experimental basis for examining how transferred representations behave as crop identity, class structure and visual diversity change across successive target domains. The findings therefore provide a reference point for future cross-crop adaptation and multi-crop disease-recognition research.

The study is limited by its reliance on image datasets and the absence of direct field deployment or real-time edge evaluation. Further validation using field-acquired imagery, broader environmental variation and additional crop domains would provide a stronger assessment of transferability and practical deployment. Future work may also investigate continual learning and adaptive domain-transfer strategies to determine whether the observed transfer behaviour can be maintained as agricultural domains expand.

## Data availability

The datasets used in this study are publicly available from their respective third-party repositories and are cited in the manuscript (Refs. 23 and 30–33). The authors do not own these datasets and therefore do not have permission to redistribute or deposit the original datasets. Access to the datasets is subject to the terms and conditions of their respective repositories.

## Declaration of generative AI and AI-assisted technologies in the manuscript preparation process

During the preparation of this work, the author used OpenAI ChatGPT (GPT-5.6 Luna) to assist with copy-editing and refinement of the author's existing manuscript text for language clarity, readability, organisation, and formatting. After using this tool, the author reviewed and edited the content as needed and takes full responsibility for the accuracy, originality, integrity, scientific content, data, analyses, and conclusions of the manuscript.